\documentclass[lettersize,journal]{IEEEtran}
\IEEEoverridecommandlockouts
\usepackage{cite}
\usepackage{amsmath,amssymb,amsfonts}
\usepackage{algorithmic}
\usepackage{algorithm}
\usepackage{comment}
\usepackage{graphicx}
\usepackage{textcomp}
\usepackage{xcolor}

\DeclareMathOperator*{\argmin}{arg\,min}
\def\BibTeX{{\rm B\kern-.05em{\sc i\kern-.025em b}\kern-.08em
    T\kern-.1667em\lower.7ex\hbox{E}\kern-.125emX}}

\begin{document}

\title{LaMI-GO:  Latent Mixture Integration for Goal-Oriented Communications Achieving High Spectrum Efficiency}
\author{\IEEEauthorblockN{Achintha Wijesinghe,~\textit{Student Member, IEEE}, Suchinthaka Wanninayaka, Weiwei Wang, 
 \\ Yu-Chieh Chao, Songyang Zhang,~\textit{Member, IEEE},   and
Zhi Ding, \textit{Fellow,~IEEE}}
\thanks{*Preprint}}
\newcommand{\model}{LaMI-GO\,}
\maketitle

\begin{abstract}
The recent rise of semantic-style communications includes
the development of 
\textit{goal-oriented communications} (GO-COMs) 
remarkably efficient multimedia information transmissions.
The concept of GO-COMS leverages advanced artificial intelligence (AI) tools to address the rising demand for bandwidth efficiency in applications, such as edge computing and Internet-of-Things (IoT). Unlike traditional communication systems focusing on source data accuracy, GO-COMs provide intelligent message delivery catering to the special needs critical to accomplishing downstream tasks at the receiver. 
In this work, we present a novel GO-COM framework, namely LaMI-GO that
utilizes emerging generative AI for better quality-of-service (QoS) with ultra-high communication efficiency. Specifically, we design our LaMI-GO system backbone based on a latent diffusion model followed by a vector-quantized generative adversarial network (VQGAN) for efficient latent embedding and information representation. 
The system trains a common-feature codebookthe receiver side. Our experimental results demonstrate substantial improvement in perceptual quality, accuracy of downstream tasks, and bandwidth consumption over the state-of-the-art GO-COM systems and establish the power of our proposed LaMI-GO communication framework.
\end{abstract}


\section{Introduction}

Future wireless networks are poised to incorporate advanced artificial intelligence (AI) techniques
and to deliver unprecedented improvement of user experiences and human-machine interactions \cite{6gintro}. 
Advanced learning technologies, including generative AI and large language models, continue to propel the 
AI revolution and change our lifestyle in an increasingly networked world. Leveraging AI and de-emphasizing
the importance of bit-level communication accuracy and capacity~\cite{shannon}, the new paradigm
of goal-oriented communication (GO-COM) makes it possible to satisfy the receiver's needs despite bit losses during communication.  Unlike conventional communication systems aimed at delivering source
data (bits) accurately,  GO-COM prioritizes addressing the receiver’s specific needs over exact bit-wise 
message reconstruction~\cite{babarosa, diffgo,diffgo+}. This new paradigm makes it possible to use
high computation intensity in exchange
for bandwidth or capacity, facilitated by continual advancements in the
semiconductor and computer industries. 
Consequently, we anticipate a design shift from classical to goal-oriented 
communication systems, where AI serves as the backbone of future-generation communication systems by 
leveraging powerful computational resources, deep learning intelligence, and immense data points.

Unlike conventional communication systems that are data-oriented, 
GO-COM systems are driven mostly by downstream tasks that
the receiver may perform upon data reception \cite{go-com1}. For example, a network of intelligent transmitters for traffic control system covering a metropolitan region 
requires no information about the exact color of each passing vehicle 
or the window shape of a background building. Without the need
for many ground-truth specifics,  one can improve
the bandwidth utility by introducing task-specific GO-COM
to remove irrelevant or private details from data communication.
As a result, goal-oriented communications aim 
to reduce bandwidth usage by not communicating 
potentially redundant information
or by obscuring private information. 

Recent GO-COM systems increasingly leverage generative AI owing to their capacity to generate information from prior distributions with minimal conditions. Among various options, diffusion-based GO-COM systems have 
appeared in works such as~\cite{babarosa,diffgo,diffgo+,bb2, diff-go-n}. The authors of \cite{bb2} presented a context-preserving image coding diffusion model; however, its transmitter sends both a segmentation and a low-resolution version of the original image, consuming significant bandwidth. Additionally, to combat channel noise and distortion,
the work in~\cite{babarosa} proposes a model that uses semantic segmentation to increase resilience against channel noise. Although this model demonstrates robustness to additive channel noise 
in source data, such a noise model is incompatible with most modern-day packet communication systems 
where channel noises and distortions lead to packet errors instead of noisy source data. 
Moreover, a pre-defined single semantic 
segmentation may omit crucial information needed for downstream tasks, such as 
missing traffic signs in images captured by remote cameras for remote-controlled robotic vehicles.

In response, the systems in~\cite{diffgo} and~\cite{diffgo+} address vital information loss 
by implementing a novel local feedback mechanism for quality assurance. This feedback also 
supports noise latent approximation to ensure consistent signal recovery at both transmitter and 
receiver ends. However, these approaches increase computational complexity and extend recovery delay
in reconstruction. Alternative approaches for GO-COM include models based on vector quantized variational auto-encoders~\cite{vqvaer}, auto-encoders~\cite{ae}, and semantic channel coding~\cite{channel}.

For better user adaptability and user privacy, 
an efficient GO-COM system is expected to allow 
receivers to carry out multiple potential downstream tasks 
with acceptable accuracy, which often benefits from training 
GO-COM modules on a very large dataset. 
To improve the GO-COM efficiency, a critical design choice for GO-COM is the backbone architecture. 
Early diffusion-based GO-COM systems initialize message reconstruction 
at the receiver with random noise, guided by manually selected conditioning, 
to exploit the diffusion model’s capacity for capturing complex distributions within the pixel domain. 
Such designs may be inefficient for practical applications. 
Take images for example, pixel domain inherently involves redundant information, 
which requires more computation and resources throughout the diffusion process. 
Furthermore, initializing the message recovery from a random noise with zero information on the 
source message introduces valid concerns regarding communication reliability. 

In contrast to these early diffusion-based approaches, traditional-style communication systems
based on deep learning often employ an auto-encoder architecture that learns a feature latent space 
capturing essential message characteristics, with reconstruction initiated directly from this 
learned latent vectors. Borrowing ideas from this approach can be advantageous. First, 
it enables a more compressed yet representative encoding of the message, 
preserving only essential features and thus reducing bandwidth consumption. 
Second, by defining a reconstruction process within this optimized latent space, 
the system achieves greater efficiency and reduces the time required for 
message reconstruction. Drawing lessons from this traditional path, it becomes evident that sharing essential 
message features with the receiver is critical. However, while GO-COM systems emphasize 
communicating the contextual meaning of a message, a balance between feature sharing and 
contextual integrity is needed to address the limitations of current GO-COM methods.

To this end, our proposed LaMI-GO framework introduces
dual emphases:  text-based diffusion
conditioning to capture message context and an auto-encoder to extract critical features. 
Inspired by vector quantization-based codebook learning~\cite{vqvae}, our approach 
constructs a quantized embedding space based on training, which is shared with the 
receiver as a one-time configuration mechanism and used as the diffusion space for LaMI-GO. 
This learned embedding space serves as a codebook, enabling efficient communication through 
codeword indexing to conserve bandwidth. Our results demonstrate that learning a quantized 
codebook on a large dataset enables our model to generalize across multiple datasets, 
a capability facilitated by our novel latent pruning strategy. 
This strategy effectively decouples the need for retraining the diffusion model, 
allowing us to leverage pre-trained models even on unseen datasets.

Our contributions are summarized as follows:

\begin{itemize}
    \item We propose a novel GO-COM system based on a quantized latent diffusion model trained over
a large dataset to eliminate manual conditioning selection. The 
text-based diffusion conditioning is supported by lightweight image captioning models.

\item We introduce a latent mixture integration strategy to generate images 
    by using a pre-trained diffusion model to avoid dedicated model
    training for specific testing datasets. In fact, we show that pre-training the backbone models on large datasets 
enables GO-COM to operate effectively on largely distributed data at communication showtime.
    \item We propose several innovative semantic compression strategies to reduce transmission bandwidth for the 
proposed latent mixture integration strategy.
    \item Through extensive evaluations, our method achieves competitive bandwidth reduction against benchmark methods and shows
    superior performance in terms of both perceptual quality metrics and downstream task-based metrics alongside minimal showtime.

\end{itemize}

\begin{figure*}[!t]
\centering
\includegraphics[width=\linewidth]{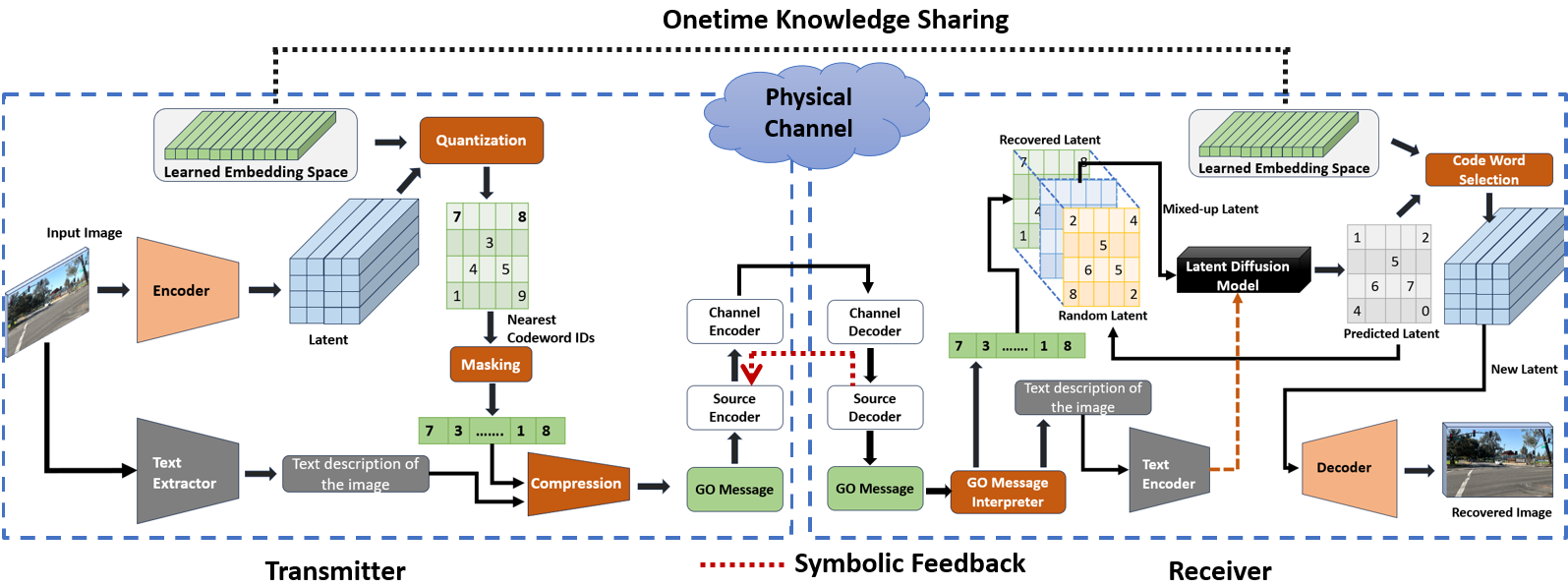}
\caption{Overall architecture of the proposed \model. Transmitter: \model has an encoder model to extract essential information of a given image. This information is then quantized using a learned dictionary and masked in the later step. Concurrently, \model extracts text information from the same image using a text extractor and compresses it along with the sequence of codeword indices. Receiver: GO, message interpreter, recovers the codeword sequence and the text;  Information is regenerated using a latent diffusion model conditioned on text embeddings with the recovered latent vectors as the input. \model proposes a latent mixture integration strategy for image recovery.
}
\label{fig_1}
\end{figure*}

The remainder of this work is organized as follows. We first provide an overview of the related works in Section \ref{sec:related}, after which our LaMI-GO system architecture and training strategies are introduced in Section \ref{sec:system} and Section \ref{sec:method}, respectively. Following the presentation of the experimental results in Section \ref{sec:results}, we summarize our work and suggest several future directions in Section \ref{sec:conclusion}.

\section{Related Work and Background} \label{sec:related}

\subsection{Overview of Goal-Oriented Communication}
To revisit several existing GO-COM frameworks, the main principle behind GO-COM is
based on intelligent information exchange without sharing all available source information with the receiver. 
In short, GO-COM needs intelligent transmitters and receivers that can benefit from AI-based joint optimization beyond earlier system models for GO-COM~\cite{R1}. 

The recent advances in generative AI have led to a paradigm shift in GO-COM. Earlier GO-COM
models applied deep learning principles. Among many 
deep learning models, the auto-encoder (AE) was one 
popular choice with end-to-end training~\cite{NR6}. However, classical AE has some known
shortcomings, including limited bandwidth savings, dataset dependability, and weaker robustness. 
VQ-VAE in~\cite{vqvaer} has mitigated some of these shortcomings. Other popular GO-COM systems considered reinforcement learning~\cite{NR11}, distributed learning~\cite{NR16}, multi-modal learning~\cite{NR7,NR12,NR13}, 
prior to the popularity of generative AI.

Several recent GO-COM designs based on generative AI such as Diff-GO~\cite{diffgo}, Diff-GO+~\cite{diffgo+}, and GESCO~\cite{babarosa} have incorporated the more advanced 
diffusion model backbone. These diffusion-inspired
GO-COM systems have shown better performances in bandwidth and perceptual quality~\cite{diffgo,diffgo+,babarosa}. In GESCO~\cite{babarosa}, the 
authors use a pre-trained diffusion model on the receiver side 
to regenerate the encoded
source signal (of an image) by letting the transmitter share the
high-level image conditions such as semantic segmentation. 
Through pre-training against additive noise channels, 
GESCO shows some robust performances against Gaussian noise.
However, the model does not provide perceptual quality or quality control in message reconstruction. 
To accommodate broader downstream tasks such as 
training a deep neural network (DNN),
Diff-GO and Diff-GO+ implement identical diffusion models at both the transmitter and receiver sides. 
For quality control,
the transmitter uses the same reconstructing process as the receiver 
in a new local generative feedback to assess reception quality by adjusting shared
conditioning information. These models aim to reduce both recovery time and 
bandwidth consumption, along 
with a higher perceptual quality, while the computation 
efficiency can be further improved.

In this work, we aim to improve diffusion-based GO-COM performance in terms of bandwidth efficiency, perceptual quality, show-time, and noise robustness. 

\subsection{Diffusion Models}

As a short review, diffusion models~\cite{diffusion} use two processes for image generation. First, a forward process adds noise iteratively to an image to transform the image into a noise space. Next, in the reverse/backward process, 
a deep neural network (DNN) is trained to successively estimate the noise added to the image at step $t$, 
given the noisy image at step $t+1$. We denote an original image by $x_{0} $ with distribution $ q(x_0) $.
Then $\{\mathbf{x_1}, \cdots,\mathbf{x_T}\}$ represent the sequence
of noisy latent representations of size identical to $\mathbf{x_0}$ that are generated 
by iterative noise addition. 

In the forward diffusion, the
posterior $q(\mathbf{x_{1:T}} |\mathbf{x_0})$ form a Markov chain that
gradually adds Gaussian noise to the given data point. Each iteration
uses a variance schedule $\{\beta_1, . . . , \beta_T\}$ as follows:
\begin{align}
   q(\mathbf{x_{1:T}}|\mathbf{x_0}) & = \prod_{t=1}^{T} q(\mathbf{x_t}|\mathbf{x_{t-1}}),\\
   q(\mathbf{x_t}|\mathbf{x_{t-1}}) & = \mathcal{N}(\mathbf{x_t};\sqrt{1-\beta_t}\mathbf{x_{t-1}},\beta_t\mathbf{I}).
    \label{eqn:fwd}
\end{align}

On the other hand, a DNN functions as a denoising auto-encoder to learn the added noise 
from each forward process to characterize the joint distribution $p_{\theta}(\mathbf{x_{0:T}})$ 
in backward diffusion. This auto-encoder is expected to learn the 
Gaussian transitions which start at $p(\mathbf{x_T}) = \mathcal{N}(\mathbf{x_T};\mathbf{0},\mathbf{I})$ 
and follows the transition below:
\begin{align}
   p_{\theta}(\mathbf{x_{0:T}}) &= p(\mathbf{x_T}) \prod_{t=1}^{T} p_{\theta}(\mathbf{x_{t-1}}|\mathbf{x_t}),\\
   p_{\theta}(\mathbf{x_{t-1}}|\mathbf{x_{t}})& = \mathcal{N}(\mathbf{x_{t-1}};\nu_{\theta}(\mathbf{x_t},t),\Sigma_{\theta}(\mathbf{x_t},t)).
    \label{eqn:bwd}
\end{align}

\subsection{VQGAN}
Vector-quantized generative adversarial networks 
(VQ-GAN)~\cite{vqgan} can
generate high-quality perceptual images based on the vector quantized variational auto-encoder 
(VQVAE)~\cite{vqvae}. In both VQVAE and VQGAN,  
encoder and decoder models are trained in a quantized space 
by learning a vector codebook to represent the essential features of the given data. In both methods, the nearest neighbor codeword within the learned codebook is used to encode an image. Finally, the decoder remaps
this quantized latent representation back to the image space. In VQGAN, a discriminator applied to the decoded image further enhances the recovery quality.

\section{Overall System Architecture} \label{sec:system}

This section introduces a novel \textit{goal-oriented} communication (GO-COM) system that utilizes a latent diffusion model in a quantized 
latent domain. Fig.~\ref{fig_1} illustrates the overall system architecture of \model. Our system design is inspired by dictionary learning as it enables models to learn image representation using the most fundamental structures. These learned representations then provide the common 
knowledge that can be leveraged to compress and represent similar but new unseen images by the model. 
For this purpose, we learn an embedding space over a large dataset. Subsequently, 
the codewords (elements of this embedding space) capture the fundamental structure of the image. 
The importance of the above representation for a communication system is twofold. First, 
this learned knowledge can be pre-shared between the 
transmitter and receiver {\em a priori} at the handshaking/configuration
stage. Second, the transmitter only needs to communicate the index value 
of each latent embedding to the receiver, instead of the full latent 
representations, to achieve high spectrum efficiency. We describe the learning process of the embedding 
space in Section~\ref{sec:method}.

\subsection{Deployment of Transmitter End }
At the transmitter, we propose to deploy an encoder to map a given image to a previously learned embedding space. For this purpose, we adopt the convolution-based encoder proposed by VQ-GAN~\cite{VQ-GAN}. Mapping images into a latent representation is paramount to capturing the most important features that describe an image. In contrast, many existing goal-oriented communication methods such as GESCO~\cite{babarosa}, Diff-GO~\cite{diffgo}, and Diff-GO+~\cite{diffgo+} bypass this step and try to reconstruct an image from a randomly sampled noise image in the original 
image space. As a result, we notice a significant performance degradation as elaborated in Section~\ref{sec:results}. Parallel to image encoding, 
we also derive \textit{text description} of the given image. This text description is generic and imitates the image caption datasets. 
For image captioning, we 
select lightweight and fast models for efficient communication such as~\cite{imagetotxt}. In this work, we compress the text description using 
Huffman or entropy coding for bandwidth reduction.  

Once the image is mapped to the embedding space, we apply latent quantization via dictionary learning. As described above, we leverage a set of pre-trained embeddings (codewords) in a codebook and map each mapped
latent embedding to the index of the nearest codeword in the learned codebook. Therefore, the latent space is consequently represented by a sequence of integer indices which form a highly compressed representation of high dimensional embeddings. In \model, the encoding achieves a factor of $16\times$ compression of the original image. We further compress the representation by selectively masking some indices of low importance. This operation helps us include the semantic aspect of the model. Finally, we propose to compress the index sequence using Huffman or entropy coding in the compression block together with the text conditions. 
The output is the derived \textit{GO message} which is then sent to the receiver using the physical channel.

\subsection{Receiver Operation and Data Regeneration }

Modern-day receivers in communication systems are equipped 
with state-of-the-art hardware, capable of efficiently processing complex computations. To leverage hardware support, \model shifts much of the computation load to the receiver end. Our design exploits the latent diffusion model Paella~\cite{paella} as the backbone generative model. 
Upon receiving the GO message, the GO message interpreter generates the received codebook index sequence and the text condition.
The text conditions are sent to a ByT5~\cite{byt5} and/or Clip~\cite{clip} encoders to recover the latent condition denoted by $\mathbf{c}$. In parallel, the recovered index sequence is also a mixture of random or predicted latent indices (in later iterations) to generate the input latent to the diffusion model. Since the diffusion process is iterative,  early iterations can be explored by mixing up the input with 
randomly generated latent embeddings;  whereas towards the end, the mixture integration process uses predicted latent embeddings.  In each iteration, the derived input and the corresponding conditions are fed to the latent diffusion for
latent prediction. Note that, unlike classical diffusion models, this new model is trained to predict the integer index of the corresponding latent embedding as shown in Fig.~\ref{fig_1}. After $T$ iterations, the 
predicted latent embedding is extracted from the codebook. 
We utilize the VQ-GAN decoder model to
map the predicted latent embedding back to the image domain.

\subsection{Noise Model}

\begin{figure}[!t]
\centering
\includegraphics[width=\linewidth]{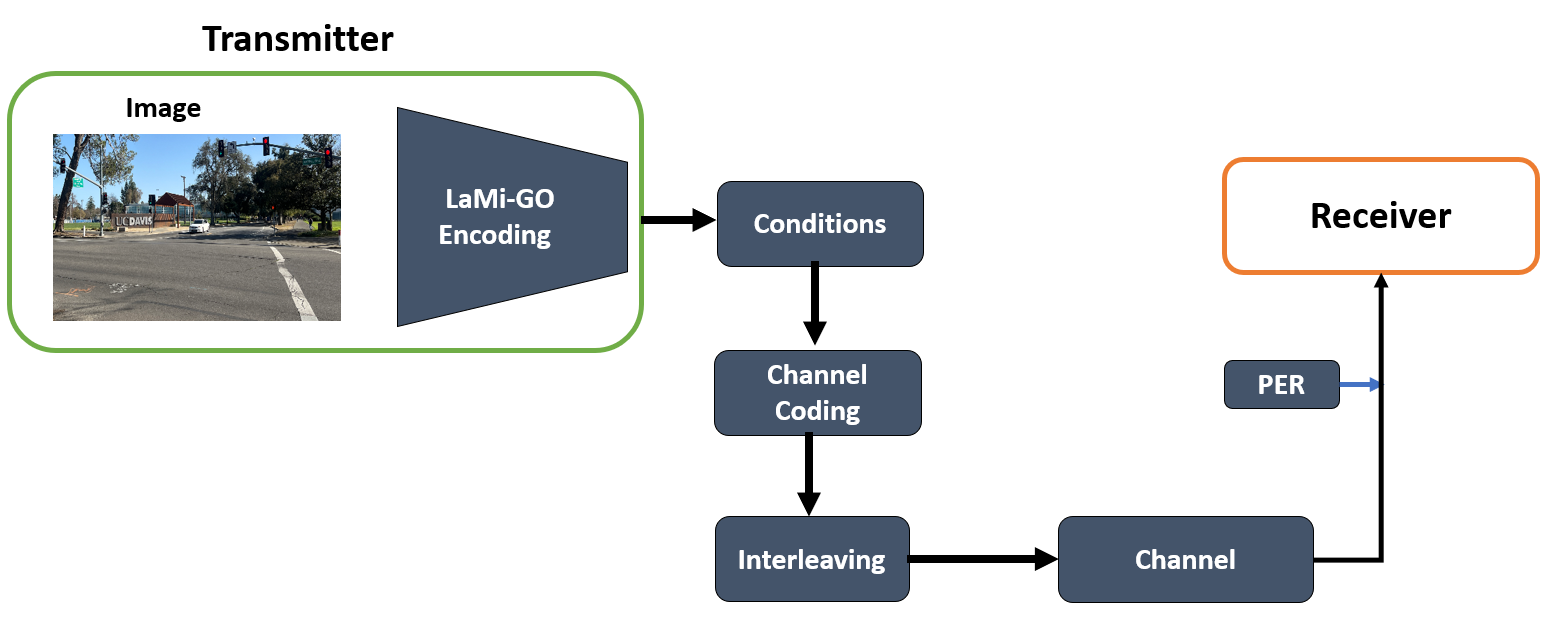}
\caption{Channel noise model. We consider digital communication with packet transmission. 
}
\label{fig_channel}
\end{figure}

Fig.~\ref{fig_channel} presents the overall channel modeling
considered in this paper. In this proposed digital communication system, an inference image is 
converted to a GO message by the LaMI-GO encoder.  
Next, the GO message is protected using channel coding. 
We propose adopting a widely used interleaving operation 
in our pipeline 
to mitigate the effects of channel impairment on the message. For empirical evaluation of the effect of the channel noise on our system, we consider several setups. In noise setup 1 and noise setup 2, we consider interleaving operation. Also, in setup 1, we consider Huffman source coding and drop any packet with errors. These dropped packets are represented as the packet error rate (PER). In setup 2, we apply bit-wise noise as bit flipping with a probability of $p$. We adopt fixed-length coding for the simulations to prevent complications from the source decoding process. Moreover, no error correction methods are included. For setup 3 and setup 4, we do not include the interleaving operation. Similar to the previous cases, setup 3 represents packet-level error, and setup 4 represents bit-level error.

\section{Detailed Structures} \label{sec:method}

This section elaborates on the details of different system blocks in LaMI-GO.  
\subsection{Latent Diffusion Backbone for LaMI-GO}
To empower our GO-COM system, we utilize a text-to-image diffusion model from the work Paella \cite{paella}. Unlike other existing GO-COM works, we introduce the diffusion process in the VQ latent space. As previously mentioned, each latent vector is one of the $K$ codewords of the latent space assuming $K$ learned codewords. Therefore, the denoising process of the noisy latent can be viewed as a classification task. Note that with this formulation, we can represent a given latent (a set of latent vectors) as a set of indices corresponding to the learned codebook.

Assume a text-conditioned diffusion model: $f_{\phi}(. |t,\mathbf{C})$, takes a 2D matrix of indices $\mathbf{I} \in \mathbb{Z}^{h' \times w'}$ and outputs a 3D tensor $\mathbf{I}_o \in \Re^{h' \times w' \times K}$  and converted to a likelihood tensor using SoftMax operation. Here, the model's learnable parameters are represented by $\phi$. 
i.e., For a given text condition $\mathbf{C}$ and a time step $t$, the diffusion model $f_{\phi}(. |t,\mathbf{C})$ denoises a 2D matrix of indices and output the likelihood of each index against $K$ codewords. Note that, the denoising process is iterative, and $t \sim \mathcal{U}(0, 1)$ represents the severity of the noise added to the 2D matrix of indices. Unlike classical diffusion models, adding noise to the index matrix represents replacing a set of indices with randomly sampled indices from the codebook. For each index in the input, the predicted index is obtained by considering the likelihood of each index by using the SoftMax function and multinomial sampling without selecting the maximum likelihood. If we assume the input index matrix to be $\mathbf{I}$ and $\Pi$ to represent multinomial sampling, the predicted latent at time $t$ can be seen as follows,

\begin{flalign}
   \mathbf{I}_p = \Pi(\mbox{softmax}(f_{\phi}(\mathbf{I}|t,\mathbf{C}))).
    \label{eqn:mn}
\end{flalign}

We direct curious readers to the work~\cite{paella} for information regarding model training and sampling.

\subsection{Task Formulation}

We model the GO-COM message generation task on the receiver side as a classification task by utilizing the Paella text-to-image diffusion model. Let $E_{\theta}(\cdot)$ and $D_{\psi}(\cdot)$ be the encoder and the decoder 
models of VQ-GAN with their respective learnable parameters $\theta$ and $\psi$. 
Given an image $\mathbf{x_0} \in \Re^{(h \times w \times c)}$ with image height $h$, width $w$, and channel size $c$,
the encoder $E_{\theta}(\cdot)$ generates
$\mathbf{z}$ as the
latent representation of $\mathbf{x_0}$,
\begin{equation} 
\mathbf{z} = E_{\theta}(\mathbf{x}_0) \in \Re^{h' \times w' \times \ell}.
\end{equation}

Note that, $h'= \lfloor h/s\rfloor$ and $w'= \lfloor w/s \rfloor$ result from the compression factor $s$ for encoding and $\ell$ is the number of latent channels. We then reshape $\mathbf{z}$ into a collection of $m=h'\times w'$ latent vectors, each has length $\ell$, to facilitate further discussion:
\begin{equation}
\label{eqn:apr1}
    \mathbf{z} \rightarrow \mathbf{z}_{\text{flat}}
=[\mathbf{z}_1,\mathbf{z}_2\cdots,\mathbf{z}_m]^T \in\Re^{m\times \ell},
\end{equation}
where, $\mathbf{z}_i\in \Re^{\ell},i=1,2,\cdots,m $ represents the latent vector at the 
$i$-th spatial position in the original $\mathbf{z}$. The latent vectors $\mathbf{z}_{\text{flat}}$ are then quantized using nearest-neighbor matching, denoted by $\mathcal{Q}(\cdot)$, based on a learned codebook $\mathcal{W} = \{\mathbf{w}_1, \mathbf{w}_2, \dots, \mathbf{w}_K\}$. Here each codeword $\mathbf{w}_i \in \Re^{\ell}$ and $K$ is the number of codebook entries. The quantization process can be expressed as:
\begin{equation}
    \hat{\mathbf{z}}_{\text{flat}} = \mathcal{Q}(\mathbf{z}_{\text{flat}})
\end{equation}
where for each latent vector $\mathbf{z}_i \in \Re^\ell$, similar to~\cite{VQ-GAN}, we search the nearest neighbor within the 
codebook: $\mathbf{w}_k\in\mathcal{W}$ to yield,
\begin{equation}
\label{eqn:apr2}
k_i=\argmin_j||\mathbf{z}_i-\mathbf{w}_j||_2^2,
\quad i=1,\;\cdots, m
\end{equation}
Once the latent vectors are quantized, they are reshaped back into their original 3D tensor form $\hat{\mathbf{z}}\in \Re^{h'\times w'\times \ell}$. This reshaped tensor is then used as input to the decoder $D_{\psi}(\cdot)$, which reconstructs the original image:
\begin{equation} 
 \hat{\mathbf{x}}_0 = D_{\psi}(\hat{\mathbf{z}}) \in \Re^{h \times w \times c}. 
\end{equation}
Note that $\hat{\mathbf{z}}$ has dimensions $(h' \times w' \times l)$. 
To communicate the high-dimensional $\mathbf{\hat{z}}$, a shared codebook provides a simple representation $I = [k_1, k_2, \dots, k_m]$ 
, which is a sequence of integer indices of the corresponding codeword from the shared codebook
and is called the ``index message''. Here each quantized latent embedding (token) is converted to a position-aware integer sequence that can be highly compressible.

In \model, we selectively mask 
some ``high redundancy indices'' (HRI) of $I$ using different policies during transmission. Our approach is to replace these HRIs with the least probable codeword index identified during training or erase them depending on the masking strategy. These masking strategies enable variable-length entropy-based coding to compress $I$ to achieve a competitive bandwidth. We explain masking strategies in detail in section~\ref{ssec:masking}. For that, we view $I$ as a 2D matrix of size $(h' \times w')$ and name it $\hat{\mathbf{I}}$. i.e.,
\begin{equation}
  \hat{\mathbf{I}} \stackrel{\triangle}{=} \text{reshape}(I,(h', w'))
\end{equation}

Next, consider the recovery of the masked HRIs on the receiver end. Even though the masked HRI is obscured on the receiver end, by imposing pre-trained dictionary selection, the recovered indices on the receiver end must belong to one of the codewords. Hence, recovering the HRI on the receiver becomes a classification problem with $K$ number of classes for codebook size $K$. The HRI recovery mechanism is not the focus of our GO-COM framework. But, if we replace these masked HRI with randomly selected learned codewords, it mimics the forward diffusion process (noising indices) of Paella. Hence, denoising these indices using Paella is a viable option. Therefore, we simply follow the method of Paella which is trained to reverse diffuse the noise (masked) indices by (1) generating a probability distribution 
for a given HRI; and (2) randomly sampling from the estimated distribution by multinomial sampling.\\ 
\noindent Remark: this approach to handling many distributions may sound complex. The power of neural networks provides a simple way of handling such distributions.

However, the generative process of Paella is unfavorable for communications. To understand 
this, let  $\pi_e$ denote the positions of replaced indices in
$\mathbf{\hat{I}}$ during the masking process
and $\pi_u$ denotes the unaffected positions. 
Let the index message $\mathbf{\hat{I}}$
after masking 
and replacement of $\pi_e$ be $\mathbf{\hat{I}}_e$. Now the receiver needs to recover the replaced codewords at positions $\pi_e$ 
in $\mathbf{\hat{I}}_e$ 
using the text-condition model $f_{\phi}(.|t_i,\mathbf{C})$ and the received
original indices at $\pi_u$
for different time stamps $t_T > \cdots > t_2 > t_1$ 
iteratively. In a communication link, 
estimating the replaced codewords at
 $\pi_e$ while retaining the codewords at
$\pi_u$ is equivalent to letting
the receiver regenerates
the message $\mathbf{\hat{I}}$
seen by the transmitter. 
As explained in previous sections, we feed $\mathbf{\hat{I}}_e$ to $f_{\phi}(.)$ with text condition $\mathbf{C}$. 

In the iterative forward/backward diffusion process in Paella, indices in both $\pi_u$ and $\pi_e$ change. Also,  for $\tau/T$ number of steps, Paella randomly replaces different fractions of (in decreasing order) indices in $\mathbf{\hat{I}}_e$ by randomly sampling the codebook. $\tau$ is a predefined constant known as the re-noising step. In each Paella iteration, codebook indices presented in $\mathbf{\hat{I}}_e$ are changed by iterative model prediction and random replacement. As a result, Paella model's 
codebook indices in $\pi_u$ would
vary in forward and backward diffusions, which
is not necessary in our communication
link since we preserve the codebook indices in $\pi_u$.
To keep the codebook indices in $\pi_u$ and to
avoid retraining the model, we introduce a latent mixture integration process in the following section.

\subsection{Latent Mixture Integration for Received Image Retrieval}
\label{ssec:mixture up}

The concept of a latent mixture integration (LaMI) is inspired by the training process of the  Paella 
model and the need for a controlled convergence in the image recovery which is not fully governed by the text conditioning. Since our proposed LaMI
process preserves codebook indices at positions in $\pi_u$, it allows the use of pre-trained VQ-GAN and the Paella diffusion model without fine-tuning or retraining. 
In Paella training, the forward diffusion randomly replaces some indices with some noise indices in each given iteration. This process is controlled by a noise ratio derived from $t$. The ratio $t$ is 
akin to the iteration step in conventional diffusion models. Therefore, during its denoising
process, Paella model varies $t$ from $t_1 = 1$ (i.e., full noise) to $t_T = 0$ 
(i.e., zero noise) which removes 
all essential features at the beginning and 
tries to reconstruct them using text conditions. However, this feature is undesirable  
for data communications because the essential meaning of a message in $\pi_u$ has 
been initially lost in the process and becomes nearly 
impossible to restore. 

This work instead focuses on establishing a new
efficient GO-COM framework. Thus, we propose a LaMI strategy that requires no training or fine-tuning of the original Paella model. 
Note that the \model receiver is 
aware of the positions $\pi_e$
of masked token identities at the transmitter.
We denote the mask matrix $\mathbf{M}$
and also define an all-ones matrix$\mathbf{1}_{h' \times w'}$ of size $h' \times w'$.
At step $i$, LaMI regenerates the codeword index
message
\begin{equation}
    \mathbf{\hat{I}}_{i} = (\mathbf{1}_{h' \times w'}-\mathbf{M})\odot\mathbf{\hat{I}}_e + \mathbf{M}\odot\mathbf{\hat{I}}_{p,i},
\end{equation}
where $\odot$ denotes the
Hadamard product 
and the replaced indices at the transmitter
is predicted (estimated) according to
\begin{equation}
\label{eqn:aprt}
\mathbf{\hat{I}}_{p,i}  =
     \begin{cases}
     \text{random sample from $\mathcal{W}$} & \text{if $i = 1$}\\
     \Pi(\mbox{softmax}(f_{\phi}(\Gamma\mathbf{(\hat{I}}_{{i-1}})|t_i,\mathbf{C})) & \text{if $i<\tau$}\\
      \Pi(\mbox{softmax}(f_{\phi}(\mathbf{\hat{I}}_{i-1}|t_i,\mathbf{C})) & \text{if $\tau < i < T $},
    \end{cases}  
\end{equation}
where $\tau$ is the re-noising step, $\mathcal{W}$ denotes the learned codebook, $softmax(.)$ denotes
SoftMax nonlinearity,
and $\Pi$  denotes multinomial sampling. Further,
$\mathbf{\hat{I}}_{p,1}$ is a random latent sample from the codebook indices space. The function $\Gamma$ represents the re-noising function defined in Paella. This function will randomly change (add noise) some portion of the given latent by initial noise latent $\hat{\mathbf{I}}_{p,1}$. The portion is determined by the noise ratio $t_i$. Note that we perform the latent mixture integration process in the integer space, which directly maps the latent space.

\begin{figure}[!t]
\centering
\includegraphics[width=\linewidth]{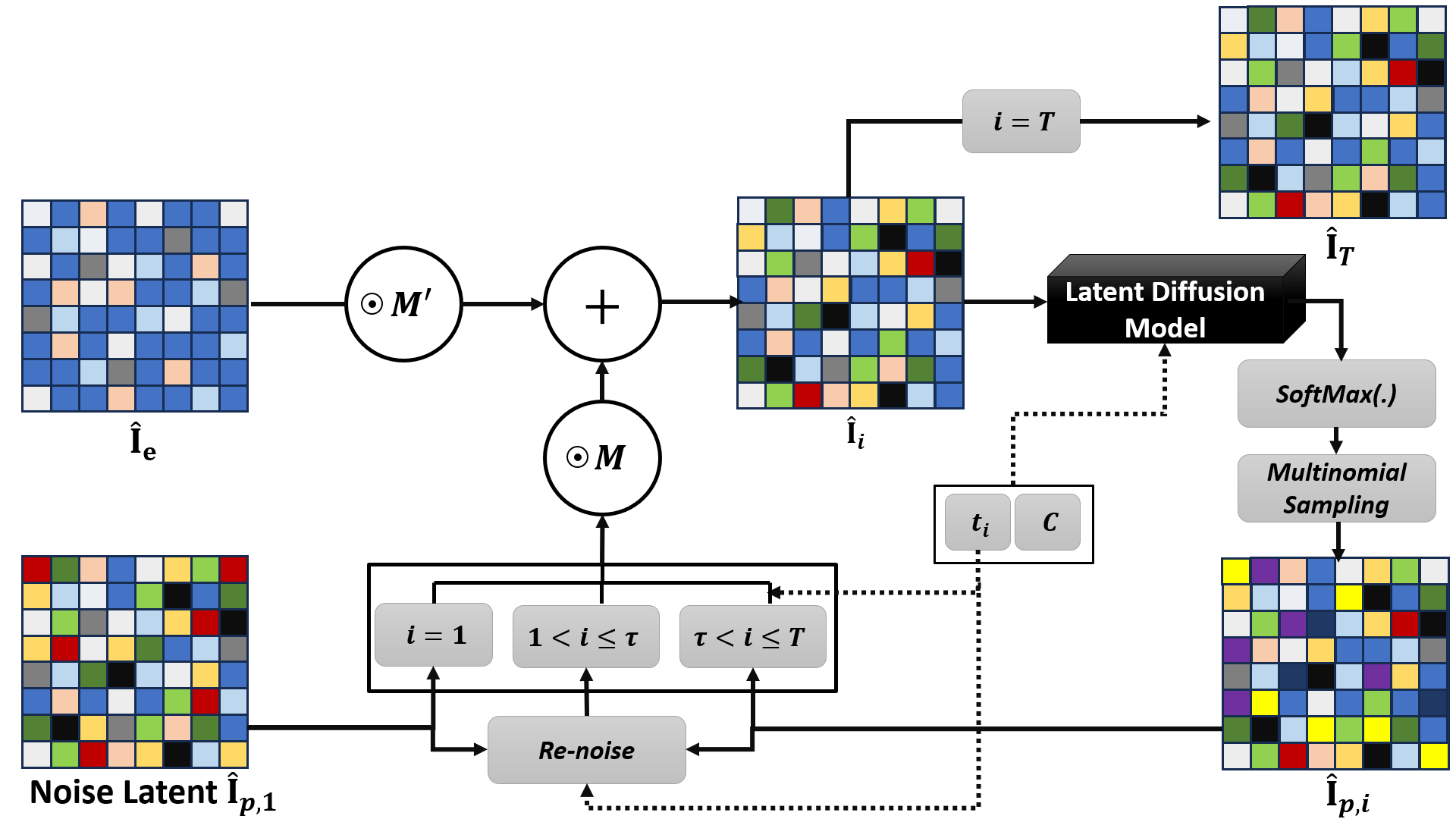}
\caption{The iterative process of latent mixture integration. Here, $M'$ represent $1_{h' \times w'} - M$.
}
\label{fig_mixture}
\end{figure}

As mentioned previously, the final image reconstruction after $T$ steps
can be written as
\begin{equation}
    \hat{\mathbf{x}}_{0} = D_{\psi}(\hat{\mathbf{z}}_{T}),
\end{equation}
where, $\hat{\mathbf{z}}_{T}$ is the corresponding latent mapping of the indices matrix $\mathbf{\hat{I}}_{T}$. Fig.~\ref{fig_mixture} illustrates the iterative process of latent mixture integration.

\subsection{Masking Policies }
\label{ssec:masking}
We now consider several possible masking policies on the transmitter side. For this work, we propose three strategies and explore more optimal policies in future works. 

\subsubsection{Pseudo-random Masking (PRM)}
First, we introduce the diffusion training process of the Paella model. As discussed above, the forward diffusion process randomly adds noise to $\mathbf{\hat{I}}$
by replacing its values 
at the masked positions 
with randomly sampled codewords from the codebook. 
Following this
strategy, we propose to simply use a pseudo-random mask (PRM)
to select the positions of $\mathbf{\hat{I}}$ 
for replacement. 
Practically, the PRM
can be easily shared
with the receiver by defining a pseudo-random
noise seed used to
generate PRM indices. This policy requires minimum
bandwidth overhead in 
communications. In our experiments, we consider masking ratio (i.e. probabilities)
of $p=\{0.25,0.35\}$ which provides an adequate tradeoff between bandwidth and the performances.
The PRM uses a mask matrix $M_r$.

\subsubsection{Pre-Determined Masking (PDM)}
 Since the encoder, decoder, and diffusion models contain convolutional neural
network (CNN) layers, it is possible to leverage neighboring data information
for recovery at the decoder. 
Thus, we consider a
pre-determined mask
(PDM) pattern for
$\mathbf{\hat{I}}$.
In the PDM strategy, we
agree on a one-time mask between the transmitter and the receiver.
Such masks may also be dynamically changed
by selecting a selected
mask pattern from
A set of pre-determined masks are generated and shared a priori.  
The transmitter only needs
to indicate to the receiver
the index of the mask within the shared mask set. 


We denote the fixed
mask used in PDM with
a masking matrix 
$\mathbf{M}_f$. Specifically, we can design a simple 2-D masking matrix for masking out $1/4$ of the tokens using the following mask,
\begin{equation}
\label{eqn:mf}
    \mathbf{M}_f (a,b)=
     \begin{cases}
      1 & \text{if $a$ and $b$ both are even }\\
      0 & \text{otherwise},
    \end{cases}  
\end{equation}
This exemplary PDM repeats a $2\times2$ pattern.

\subsubsection{Entropy Based Masking (EBM)}
Lastly, we present an entropy-based strategy for masking the $\mathbf{\hat{I}}$. As presented in Algorithm~\ref{alg:alg1}, we calculate the distance between the encoded latent 
information
and the learned embeddings at step 
Eq.~(\ref{eqn:apr1}) using the following equation for each latent vector at $(a,b)$ of $E_{\theta}(\mathbf{x_0})$  :
\begin{equation}
    \label{eqn:distance}
   d_{(a,b)} = \min_k||E_{\theta}(\mathbf{x_0})_{(a,b)} - \mathbf{w_k}||,
\end{equation}
where we denote each codeword of learned embedding space as $\mathbf{w}_k$.
Next, the mask matrix selects the
indices of the codewords with a lower distance than a
masking threshold $\eta$. 

If the set representing these indices is $D$, the EBM uses a masking
matrix $\mathbf{M}_e$ defined by
\begin{equation}
\label{eqn:md}
    \mathbf{M}_e (a,b) = 
     \begin{cases}
      1 & \text{if the coordinate pair $(a,b) \in D$ }\\
      0 & \text{otherwise},
    \end{cases}  
\end{equation}
Note that the distance $d_{(a,b)}$ is directly related to the entropy of
information loss. EBM aims to drop the quantized latent entries of lower entropy. 
On the receiver side, we utilize the conditions and prior knowledge of the diffusion model to mitigate the entropy loss at the EBM indices. 
Since $\mathbf{M}_e$ is message-dependent, the receiver needs additional information from
the transmitter to acquire the masking matrix. 
We discuss the effect of selecting the best
$n$ candidate indices for masking in Section~\ref{sec:results}. 

\subsection{Overview of Post-Training Showtime }
\label{ssec:algo}
\begin{figure}[!t]
\centering
\includegraphics[width=\linewidth]{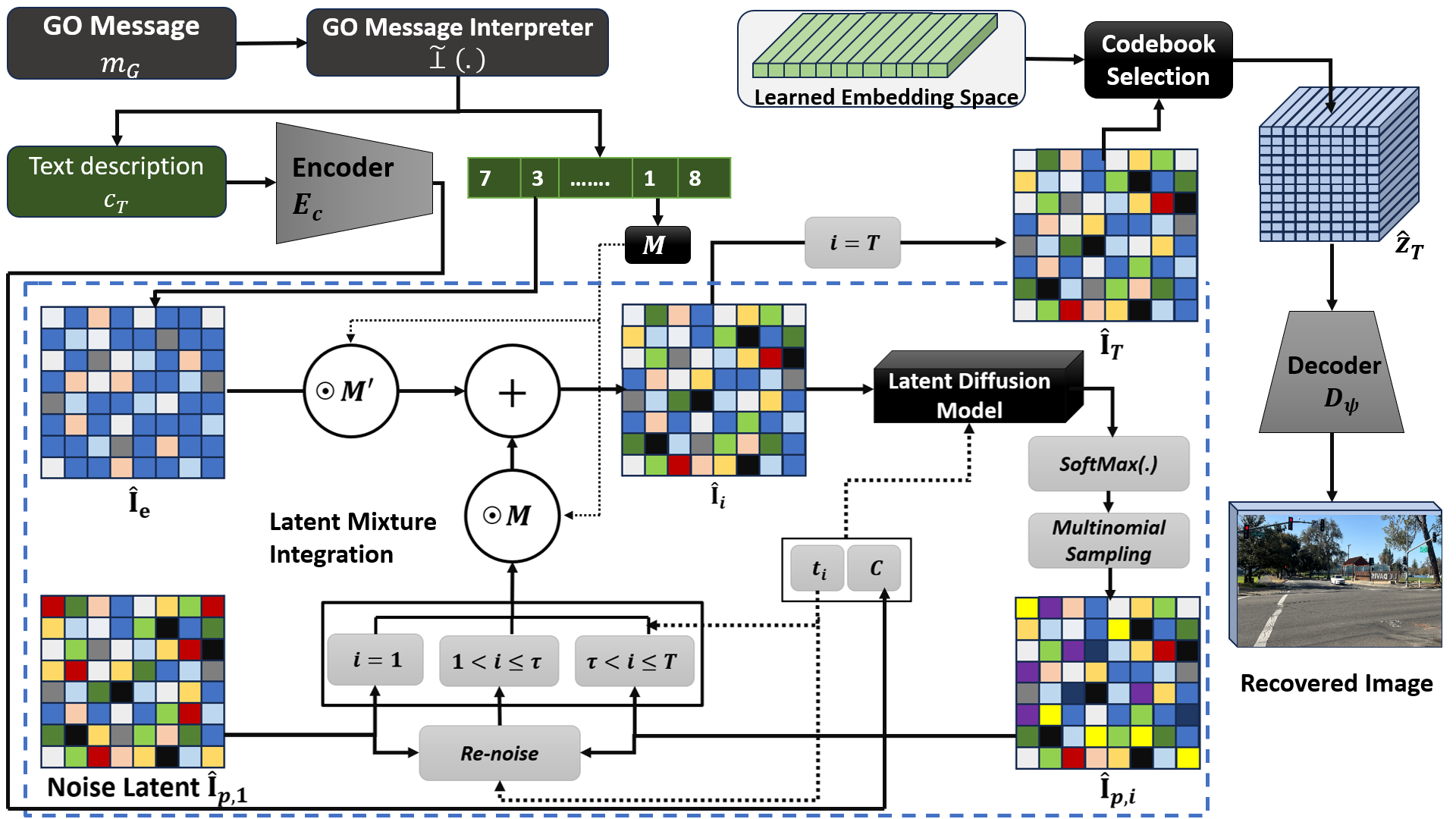}
\caption{The showtime on the receiver end.
}
\label{fig_receiver}
\end{figure}

Once the trained model is deployed,  we enter a ``showtime" period to test the
communication efficiency and efficacy. The receiver side of Fig.~\ref{fig_receiver} visualizes this process. During showtime, the GO message interpreter at the receiver  the GO message 
to generate the shared indices and semantic conditions for the latent diffusion model. 
Using a parallel path, the receiver identifies the mask according to the
received masking matrix information from the transmitter. 
Once the received conditions and the shared masked latent are ready, 
random noise vectors replace the masked latent entries by using our proposed 
latent mixture integration strategy, starting with a random latent. 
After $\tau$ steps, we replace the random latent entries with the predicted latent entries
and continue the mixture integration process to complete $T$ overall steps. Upon the completion of the diffusion iterations, the predicted latent entries are respectively matched with their
closest neighbors in the shared codebook to derive the final latent representation
for
the decoder to get the image domain representation. This reconstructed 
image then can be used for the different services on the receiver side.

We summarize the method into two algorithms as presented below in Algorithm~\ref{alg:alg1} and Algorithm~\ref{alg2:al2}.

\begin{algorithm}[htb]
\caption{\model \; Showtime at the Transmitter}
\label{alg:alg1}
\begin{algorithmic}
\STATE {\textbf{Input:} Data sample: $\mathbf{x}_0$ }
\STATE {\quad \quad \quad  Masking policy: $\mathcal{M}$  }
\STATE {\textbf{Initiate:}  Pre-trained codebook: $\Psi$ }
\STATE {\quad \quad \quad Pre-trained encoder: $E_{\theta}(.)$ }
\STATE {\quad \quad \quad Pre-trained text Encoder: $E_T(.)$ }
\STATE $\mathbf{z} \gets E_{\theta}(\mathbf{x}_0)$ \# Encode the image
\STATE  $\mathbf{\hat{z}} \gets \mathcal{Q}(\mathbf{z})$ \# Use eq.~\ref{eqn:apr1} and~\ref{eqn:apr2}
\STATE {Mask: $M \gets$ masking indices using $\mathcal{M}$}
\STATE $\hat{\mathbf{I}}_e \gets M(\hat{\mathbf{I}})$
\STATE {Text condition: $c_T \gets E_T(\mathbf{x}_0)$}
\STATE {Compressed message: $m_c \gets $compress($\hat{\mathbf{I}}_e,c_T)$}
\STATE {\textsc{SEND }} ($m_G = m_c$) with GO-Message
\end{algorithmic}
\end{algorithm}

\begin{algorithm}[htb]
\caption{\model \; Showtime at the Receiver}
\begin{algorithmic}
\STATE {\textbf{Input:} GO message: $m_G$ }
\STATE {\quad \quad \quad  Masking policy: $\mathcal{M}$  }
\STATE {\quad \quad \quad  Noising steps: $\tau$  }
\STATE {\quad \quad \quad  Denoising steps: $T$  }
\STATE {\textbf{Initiate:}  Pre-trained codebook: $\Psi$ }
\STATE {\quad \quad \quad Pre-trained decoder: $D_{\psi}(.)$ }
\STATE {\quad \quad \quad Pre-trained diffusion model: $f_{\phi}(.)$ }
\STATE {\quad \quad \quad Pre-trained text Encoder: $E_c(.)$ }
\STATE {\quad \quad \quad GO message interpreter: $\mathcal{I}(.)$ }
\STATE{($\hat{\mathbf{I}}_e,c_T) \gets \mathcal{I}(m_G)$}
\STATE{$M \gets $ decode mask using $\hat{\mathbf{I}}_e$ }
\STATE{Generate conditions for diffusion model: $\mathbf{C} \gets E_c(c_T)$}
\STATE{$\hat{\mathbf{I}}_{p,1} \gets $ random indices latent }
\STATE{$\hat{\mathbf{I}}_{1} = (\mathbf{1}_{h' \times w'}-M)\odot\hat{\mathbf{I}}_e + M\odot\hat{\mathbf{I}}_{p,1}$}


\STATE {\textbf{For $i$ in range $2: T $}:}
\STATE {\quad \quad \quad  \textbf{If $t < \tau$ :}  }
\STATE {\quad \quad \quad \quad $\hat{\mathbf{I}}_{p,i} \gets $ $\Pi(\mbox{softmax}(f_{\phi}(\Gamma\mathbf{(\hat{I}}_{{i-1}})|t_i,\mathbf{C}))$  }
\STATE {\quad \quad \quad  \textbf{Else :}  }

\STATE {\quad \quad \quad \quad  $\hat{\mathbf{I}}_{p,i} \gets \Pi(\mbox{softmax}(f_{\phi}(\mathbf{\hat{I}}_{{i-1}}|t_i,\mathbf{C}))$ }
\STATE{\quad \quad \quad $\hat{\mathbf{I}}_{i} = (\mathbf{1}_{a \times b}-M)\odot\hat{\mathbf{I}}_e + M\odot\hat{\mathbf{I}}_{p,i}$}
\STATE{$\hat{\mathbf{z}}_{T} \gets $ get latent from 
 $\hat{\mathbf{I}}_{T}$}
\STATE {Image recovery: $\mathbf{\hat{x}_0} \gets D_{\psi}(\hat{\mathbf{z}}_{T})$}
\end{algorithmic}
\label{alg2:al2}
\end{algorithm}

\section{Experimental Results} \label{sec:results}
This section provides new empirical results obtained after rigorously evaluating our proposed method. Our extensive evaluations cover several datasets, including Cityscapes dataset~\cite{cityscape}, which is popular for traffic management and autonomous driving-related tasks with 35 classes,  as well as the Flickr~\cite{flickr}, COCO-Stuff~\cite{coco-stuff}, and COCO~\cite{coco} datasets which are widely used for tasks such as object detection, segmentation, and image captioning. In the \model framework, we consider images of size $256\times512$ from Cityscapes and $256\times256$ for other datasets. We follow different evaluation strategies and metrics. 
To evaluate the image recovery quality, we employ perceptual quality metrics such as FID~\cite{FID} and LPIPS~\cite{LPIPS} alongside downstream tasks such as depth estimation and object detection. Our experiments only require a single A100 GPU with 80GB VRAM and additional CPU support. For the evaluation, we get the pre-trained Paella model trained on the LAION~\cite{lion} dataset and avoid fine-tuning or training.

\subsection{Reconstruction Quality and Overall Performance}

We first illustrate the semantic reconstruction power of \model over existing semantic/GO-COM works: Diff-GO+~\cite{diffgo+}, GESCO~\cite{babarosa}, Diff-GO~\cite{diffgo} and its variants RN (Random Noise) and OD (Original Diffusion)~\cite{diffgo}, and other semantic image syntheses models: SPADE~\cite{SPADE}, CC-FPSE~\cite{CC-FPSE}, SMIS~\cite{SMIS}, OASIS~\cite{OASIS}, and SDM~\cite{SDM}. Table~\ref{tab:table1} summarizes their results over the Cityscapes dataset. All three semantic communication methods and their variants use pixel domain diffusion models as the backbone and semantic segmentation map and edge map as the model conditions. We deliver LPIPS and FID scores with the required denoising steps. The comparison shows the superior performance of \model over other semantic generation and communication proposals. The performance for all three measures exhibits substantial improvement. One notable aspect of these results is the lower number of diffusion steps needed for \model compared with other diffusion-based semantic communication benchmarks. Because low latency and low computation complexity are critical to communication network services,  fewer diffusion steps exponentially reduce the receiver decoding time. 

We extend the test performance of \model against contemporary generative models based on generative adversarial networks (GANs), 
diffusion models, and auto-regressive models conditioned on \textit{texts}. Table~\ref{tab:tabletext} demonstrates that the proposed \model  model is flexible, competitive, and better with increasing computation at larger re-noising and reconstruction steps
$\tau $ and $T$.

\begin{table}[!t]
\caption{Semantic similarity of the reconstructed images with respect to LPIPS and FID scores along with the required denoising steps for the Cityscape dataset. The models evaluated under the same conditions are represented by {\textdagger}. } 
\label{tab:table1}
\centering
\begin{tabular}{|c||c|c|c|c|}
\hline
Method & LPIPS$\downarrow$ & FID$\downarrow$ & Steps$\downarrow$   \\
\hline
SPADE~\cite{SPADE}& 0.546 & 103.24 & NA\\
\hline
CC-FPSE~\cite{CC-FPSE}& 0.546 & 245.9& NA\\
\hline
SMIS~\cite{SMIS}& 0.546 & 87.58& NA\\
\hline
OASIS~\cite{OASIS}& 0.561 & 104.03& NA\\
\hline
SDM~\cite{SDM}& 0.549 & 98.99& NA\\
\hline
OD\textdagger& 0.2191 & 55.85& 1000\\
\hline
GESCO\textdagger & 0.591 & 83.74& 1000\\
\hline
RN\textdagger &  0.3448 & 96.409& 1000\\
\hline
Diff-GO\textdagger (n=20)& 0.3206 & 74.09& 1000\\
\hline
Diff-GO\textdagger (n=50)& 0.2697 & 72.95& 1000\\
\hline
Diff-GO\textdagger (n=100)& 0.2450 & 68.59& 1000\\
\hline
Diff-Go+\textdagger (W = 64, L = 32) & 0.2231 & 60.93& 1000 \\
\hline
Diff-Go+\textdagger (W = 128, L = 32) & 0.2126 & 58.20 & 1000\\
\hline
\model\textdagger & \textbf{0.1327} & \textbf{29.84} & 8\\
\hline
VQGAN\textdagger & \textbf{0.056} & \textbf{11.21} & -\\
\hline
\end{tabular}
\end{table}

\begin{table}[!t]
\caption{Semantic similarity of the reconstructed images with respect to FID scores on the COCO dataset ( 40k images) against existing generative methods uses text conditions. The models evaluated under the same conditions are represented by {\textdagger}. Here, we have used the fixed policy for \model training.} 
\label{tab:tabletext}
\centering
\begin{tabular}{|c||c|}
\hline
\textbf{Method} &  \textbf{FID}$\downarrow$    \\
\hline
DMGAN~\cite{DMGAN}& 32.64\\
\hline
XMCGAN~\cite{XMCGAN}& 50.08\\
\hline
DFGAN~\cite{DFGAN}& 21.42\\
\hline
SSA-GAN~\cite{SSA-GAN}& 19.37\\
\hline
DSE-GAN~\cite{DSE-GAN}& 15.30\\
\hline
VQ-Diffusion~\cite{VQ-Diffusion}& 19.75\\
\hline
VQ-GAN~\cite{VQ-GAN} & 22.28\\
\hline
Stackformer~\cite{Stackformer} & 10.08\\
\hline
\model\textdagger (28/32)& 10.26\\
\hline
\model\textdagger (194/200)& \textbf{9.50}\\

\hline
\end{tabular}
\end{table}

\subsection{Masking Policy Effects}

\subsubsection{Perceptual Comparison}

To better evaluate the performance of our model, we compare the three different masking policies, PRM, PDM, and EBM, against the perceptual quality of the received image. Table~\ref{table:multicolumn} presents the results from testing different policies across various datasets with different configurations of noising/denoising steps. 

From Table~\ref{table:multicolumn},  we see that PRM and PDM policies exhibit comparable performance across all datasets except for the Cityscape dataset. For Cityscapes, EBM demonstrates substantial improvement. We understand the reason for the additional benefits of applying entropy-based masking in EBM: 
For the Cityscape dataset, EBM removes and then reconstructs the most common and
easy-to-reconstruct image features such as large 
buildings and road surfaces. However, EMB on the remaining datasets shows lower FID scores. This is due to the more complex nature of the datasets. 
The Flickr and COCO-Stuff images contain a large number of objects in a single frame that are not seen in other regions within the image. Thus, masking out many latent
features from the same region can lead to 
noticeable reconstruction errors. We shall further analyze the entropy-based masking (EBM) policy in the following section. On the other hand, we see an improvement in the LPIPS score by EBM for all tested datasets
unlike for FID scores.  We can view LPIPS and FID as two contracting downstream tasks. 
The conclusion is that a good choice of masking policy depends on the typical 
image features within a dataset 
as well as on downstream task measures.

\subsubsection{Analysis of EBM Policy}
\begin{figure}[t]
\centering
\begin{minipage}[t]{0.32\linewidth}
\centering
\includegraphics[width=1\columnwidth]{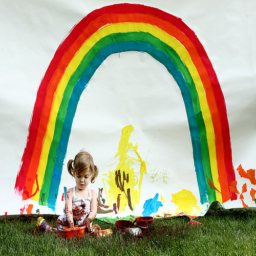}
\caption{Original image.}
\label{fig:orig}
\end{minipage}
\hfill
\begin{minipage}[t]{0.32\linewidth}
\centering
\includegraphics[width=1\columnwidth]{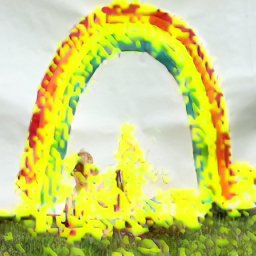}
\caption{Focus of the masked indices (in yellow) for selecting most-distanced indices.} \label{fig:h}
\end{minipage}
\hfill
\begin{minipage}[t]{0.32\linewidth}
\centering
\includegraphics[width=1\columnwidth]{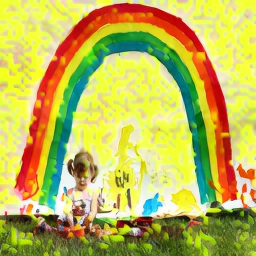}
\caption{Focus of the masked indices (in yellow) for selecting the least-distanced indices.}
\label{fig:l}
\end{minipage}

\end{figure}

EBM policy compares latent feature distance with a selection choice of either the lower distance indices or higher distance indices for masking. 
Lower distance represents the latent features closer to the learned codebook and likely captures the most common features of the dataset. On the contrary,  the higher distance represents a gap between the encoded embeddings and the learned embeddings. We believe these indices represent unique features of the source images during
the test phase (showtime).  From Fig.~\ref{fig:orig} to Fig.~\ref{fig:l}, we illustrate the original image and projection of masking indices in the pixel domain for the highest distance versus the lowest distance. Analyzing Fig.~\ref{fig:h} and Fig.~\ref{fig:l}, it is evident that the highest distance policy focuses on masking the critical and unique objects for 
the source image. Conversely, Fig.~\ref{fig:l} shows that the lowest distanced policy focuses on masking more common features with smaller variations.
Thus, if the goal of data recovery is to achieve better perceptual quality or if the goal is sensitive to many visible objects, then low-distance-based masking is preferred.  We see this phenomenon from Table~\ref{tab:tablehl}. Unlike the random or pre-determined methods, one downside of EBM in general is that its masking area may not be evenly distributed. Its masking may affect large patches of the given object. As a result, we see from Table~\ref{tab:tablehl} and Table~\ref{table:multicolumn} that the FID values of the EBM masking are generally higher.

\begin{table}[!t]
\caption{Perceptual quality of methods: highest-distanced and lowest-distanced masking. In all the cases we mask 1024 indices.} 
\label{tab:tablehl}
\centering
\begin{tabular}{|c||c|c|c|}
\hline
\textbf{Dataset} &  \textbf{Method} & \textbf{FID} & \textbf{Steps}    \\
\hline
Cityscapes & lowest-distanced & \textbf{18.44} & 4/8 \\
\hline
Cityscapes & highest-distanced & 36.98 & 4/8 \\
\hline
Flickr & lowest-distanced & \textbf{58.65} & 4/8 \\
\hline
Flickr & highest-distanced & 126.66 & 4/8 \\
\hline
\end{tabular}
\end{table}

\subsection{Impact of Masking Probability}

\begin{figure}[t]
\centering
\begin{minipage}[t]{0.9\linewidth}
\centering
\includegraphics[width=1\columnwidth]{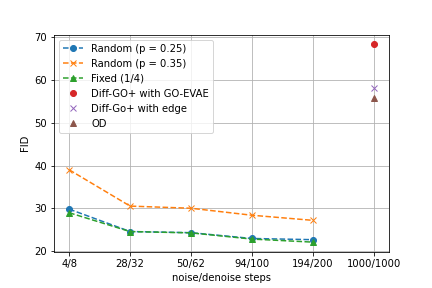}
\caption{Perceptual quality evaluation using FID for different semantic communication methods against noise/denoise steps. In the graph, we have considered different probabilities of masking on the Cityscape dataset. The x-axis represents the noise/denoise $(\tau / T)$ steps.}
\label{fig:30}
\end{minipage}

\begin{minipage}[t]{0.9\linewidth}
\centering
\includegraphics[width=1\columnwidth]{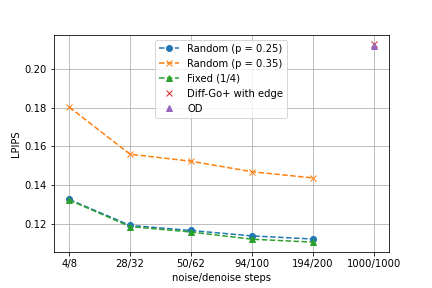}
\caption{Perceptual quality evaluation using LPIPS for different semantic communication methods against noise/denoise steps. In the graph, we have considered different probabilities of masking the Cityscape dataset. The x-axis represents the noise/denoise $(\tau / T)$ steps.}
\label{fig:100}
\end{minipage}
\end{figure}

We now investigate the effect of the sampling probability
or ratio $p$ on PRM in comparison with PDM and existing semantic communication methods. Our evaluation employs FID and LPIPS as a contextual performance metric of the recovered image on the receiver end. From Fig.~\ref{fig:30} and Fig.~\ref{fig:100}, we see that increasing masking probability $p$ from 0.25 to 0.35 lowers the perceptual quality of image recovery in the Cityscape dataset. The performance reduction at $p=0.35$ is noticeable and natural since the entropy in the received data is high. Also, the PRM model has limited capabilities to combat higher distortion rates without fine-tuning the diffusion model. As we increase $\tau$ and $T$ to allow higher delay and computation complexity, we witness the same trend of improved reduction in FID and LPIPS scores. Importantly, the performance of \model is significantly better than the previously invented semantic communication models even at the relatively higher masking ratio of $p=0.35$.
\label{ssec:noise_d}

\begin{table*}[h!]
\centering
\caption{Comparison of different policies on different datasets. }
\begin{tabular}{|c|c|c|c|c|c|c|c|c|c|}
\hline
\multicolumn{2}{|c|}{\textbf{Dataset}} &\multicolumn{2}{|c|}{\textbf{Cityscapes}} &\multicolumn{2}{|c|}{\textbf{Flickr}} & \multicolumn{2}{c|}{\textbf{COCO-Stuff}}&\textbf{noising steps ($\tau$)/} \\
\cline{1-8}
\textbf{Policy} & \textbf{Probability $(p)$} & \textbf{LPIPS} & \textbf{FID} & \textbf{LPIPS} & \textbf{FID} & \textbf{LPIPS} & \textbf{FID}  & \textbf{Denoising steps ($T$)} \\
\hline
 & & 0.1193& 24.62 &0.159 & 35.62 & 0.135 & 10.37& 28/32 \\
\cline{3-9}
Random & 0.25& 0.1167 & 24.35& 0.155 & 34.87& 0.155 & 10.04& 50/62 \\
\cline{3-9}
 & & 0.1139 & 23.00 & 0.154 & 34.77 & 0.130 & 9.75& 94/100\\
\cline{3-9}
 & & 0.1123&22.72&0.154 & 34.63 & 0.129 & 9.61 & 194/200 \\ 
\hline
 & & 0.1187&24.61&0.156 & 35.32 & 0.132 &10.20 & 28/32 \\
\cline{3-9}
Fixed & 1/4& 0.1159&24.31& 0.153&34.22 & 0.129& 9.92& 50/62 \\
\cline{3-9}
 & & 0.1122 &22.86&0.151  & 34.15 & 0.126 &9.62& 94/100\\
\cline{3-9}
 & & 0.1107&22.15&0.150 &\textbf{34.43}  & 0.126 &\textbf{9.50}  & 194/200 \\ 
\hline
 & & 0.1058&21.62&0.1510&54.15&0.126&29.31 & 28/32 \\
\cline{3-9}
Distance-based & 0.25&0.1038 &21.60&0.1477 &52.16&0.123&28.56& 50/62 \\
\cline{3-9}
 & & 0.1020 &20.52 &0.1466&52.01&0.122&28.13& 94/100\\
\cline{3-9}
 & &\textbf{0.1010} & \textbf{20.20}&\textbf{0.1461}&51.79&\textbf{0.121}&27.95& 194/200 \\ 
\hline

\end{tabular}
\label{table:multicolumn}
\end{table*}

\subsection{Evaluation on Downstream Tasks }

To assess other potential downstream task performance at the receiver end,  we further evaluate the recovered image quality against two popular downstream tasks. First, as presented in Table~\ref{tab:tabledepth}, we use depth estimation as a receiver downstream task on the recovered images from the Cityscape dataset. For comparison, we evaluate Diff-GO+ under the same test setup. From the results in Table~\ref{tab:tabledepth}, we see that the \model outperforms Diff-Go+. Moreover, there is a noticeable improvement in depth estimation accuracy for larger noise/denoise steps. These results further demonstrate that the increased diffusion steps enhance the model performance albeit at the expense of higher computation load and delay. Next, we present the object detection results of the \model using 3 different datasets in Table~\ref{tab:tableobject} in terms of the popular mAP score. For all 3 tested datasets,  the \model delivers competitive and satisfactory mAP scores. Similarly, we see that the results from both COCO-Stuff and Flickr datasets show improved image recovery quality owing to larger noise/de-noise steps. Note that because the model checkpoint is pre-trained for the COCO dataset, there is a comparatively low mAP when testing on the Flickr dataset. 

\begin{table}[!h]
\caption{Depth estimation on the Cityscapes dataset. For depth estimation, we employ the work~\cite{depthestimation}. For \model, we use random policy. } 
\label{tab:tabledepth}
\centering
\begin{tabular}{|c||c|}
\hline
\textbf{Method} &  \textbf{RMSE}$\downarrow$    \\
\hline
\model(4/8)& 5.95\\
\hline
\model(28/32)& 5.73\\
\hline
\model(50/62)& 5.72\\
\hline
\model(194/200)& \textbf{5.59}\\
\hline
Diff-GO+(1000/1000)& 6.58\\
\hline
Diff-GO+(500/500)& 6.55\\
\hline
Diff-GO+(200/200) & 6.65\\
\hline
Diff-GO+(100/100) & 6.55\\
\hline
\end{tabular}
\end{table}

\begin{table}[!t]
\caption{Object detection on the receiver end. We have used DETRs~\cite{detrs} as the object detection model. We perform object detection on the original image and the recovered image and present mAP. Here, random masking is used for image recovery and the evaluation is on a subset of each dataset: COCO-Stuff (5k images), COCO (2k images), and Flickr (2.5k images) } 
\label{tab:tableobject}
\centering
\begin{tabular}{|c||c|c|}
\hline
\textbf{Dataset} & \textbf{Steps} & \textbf{mAP }(\%)\\
\hline
& 28/32 & 63.50\\
\cline{2-3}
COCO-Stuff & 94/100 & 63.77\\
\cline{2-3}
& 194/200 & 64.21\\
\hline
COCO& 194/200 & 64.99\\
\hline
& 4/8 & 44.54\\
\cline{2-3}
Flickr& 28/32 & 44.72\\
\cline{2-3}
 & 94/100 & 46.85\\
\hline

\hline
\end{tabular}
\end{table}

\subsection{Bandwidth Consumption}

In this section, we evaluate the rate requirement for the \model as a measure of bandwidth consumption and compare it against other existing semantic communication methods, classical JPEG, and JPEG-2000. The experiments are based on the Cityscape dataset. To ensure a fair comparison, we align all methods under test to operate in a similar bandwidth range of consumption. We then benchmark the model performances based on the perceptual quality of test images and computation complexity measured by diffusion steps. From Table~\ref{tab:tableBW}, it is clear that the \model demonstrates superior performances in all the metrics. Another key highlight of the presented results is that the \model surpasses the performance of both classical JPEG and JPEG-2000.

\begin{table}[!t]
\caption{Bandwidth analysis for some of the existing semantic communication frameworks and comparison against the \model. These values are averaged across the Cityscape dataset.  } 
\label{tab:tableBW}
\centering
\begin{tabular}{|c||c|c|c|c|}
\hline
\textbf{Method}& \textbf{Bandwidth} (KB) & \textbf{FID}$\downarrow$  & \textbf{Steps}$\downarrow$   \\
\hline
Original image& 187.63  & - & -\\
\hline
GESCO& 14.52 & 83.75& 1000/1000\\
\hline
Diff-GO+(n=1024,L=1024)& 13.5 & 68.25& 1000/1000\\

+GO-EVAE(n=1024,L=16)&  & & \\
\hline
OD & 1994.41 & 55.85& 1000/1000\\
\hline
Diff-GO+(n=128,L=32)& 87.04 & 58.20& 1000/1000\\
\hline
JPEG (quality = 40) & 10.28& 35.72& NA\\
\hline
JPEG-2000 (quality = 35) & 10.94& 65.10& NA\\
\hline
\model ($M_r$ with $p=0.35$) &  \textbf{8.50} & 27.23& 194/200\\
\hline
\model ($M_r$ with $p=0.25$) &  9.39 & \textbf{22.72}& 194/200\\
\hline
\end{tabular}
\end{table}

\subsection{Performance under Channel Noise and Packet Loss }

In this section, we examine the performance of our model against channel noise in a practical communication link setup. Many deep learning-based communication systems view the channel noise effect
as directly additive to transmitted data and apply implicit training to those models. However, such a channel noise model assumes no error protection and is incompatible with the dominant packet-switched 
modern data networks in practice. Thus, we consider the channel noise effects by examining our model performance under packet loss at the receiver. Practical networks determine lost packets utilizing cyclic redundancy check (CRC) code to assess the received packet integrity. Block interleaves are commonly used to distribute lost packet
bits across multiple data packets. 
Through interleaving, packet losses can be modeled as
binary erasure channel \cite{erasure_channel}. 
Consider the random masking strategy (PRM) with $p=0.25$ for the following experiments. We leave rigorous channel noise analysis and handling as future works. In the following sections, we simply analyze the robustness of the \model against packet loss.

\begin{figure}[h]
\centering
\includegraphics[width=\linewidth]{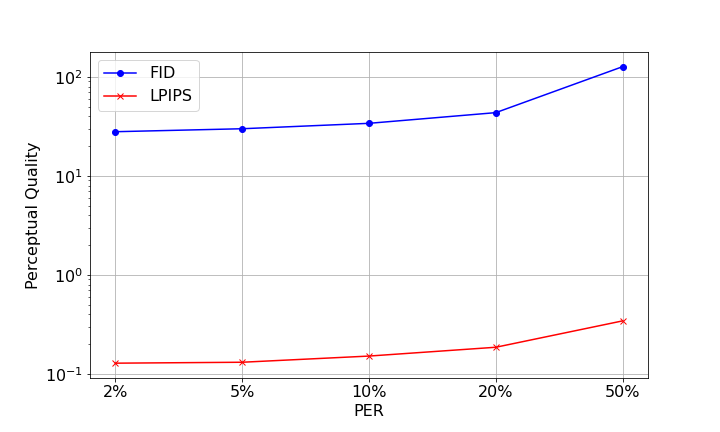}
\caption{Variation of the perceptual quality on the receiver side in terms of FID and LPIPS for different levels of packet loss.
}
\label{fig_setup1}
\end{figure}
We start with the setup 1. In this setup, we simulate our proposed communication system with all the proposed design choices against packet loss. While packet loss is a popular noise model for image and video transmission~\cite{packet1, packet2}, it simulates a digital communication system. Thus in our assessment, for each random packet loss with a packet error rate PER, the randomly lost packets 
and their payload bits are erased in our experiments.  As mentioned in the previous sections we introduce block interleaving with no channel coding at this stage. Fig.~\ref{fig_setup1} presents the perceptual quality offered by \model for different packet error rates. Note that packets contain shared indices and text conditions. Analyzing the figure shows robust performance against packet loss with good perceptual quality up to $20\%$ of the packet loss.

\begin{figure}[h]
\centering
\includegraphics[width=\linewidth]{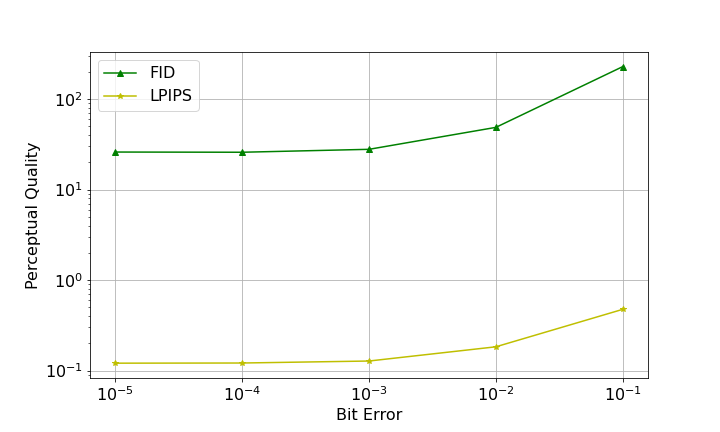}
\caption{Variation of the perceptual quality on the receiver side in terms of FID and LPIPS for different levels of bit error using interleaving.
}
\label{fig_setup2}
\end{figure} 

Next, we present the noise setup 2. In this setup, we consider bit-wise error, when block interleaving is present. We model the bit error as a flip in bits with fixed-length coding. The reason for fixed-length coding is due to the sensitivity of Huffman coding for bit-level noise which limits the simulation implementations. As portrayed in Fig.~\ref{fig_setup2}, performance of \model up to $10^{-2}$ shows acceptable perception. In fact, for practical bit noise levels such as $10^{-5}$, the model performance is very robust.

\begin{figure}[h]
\centering
\includegraphics[width=\linewidth]{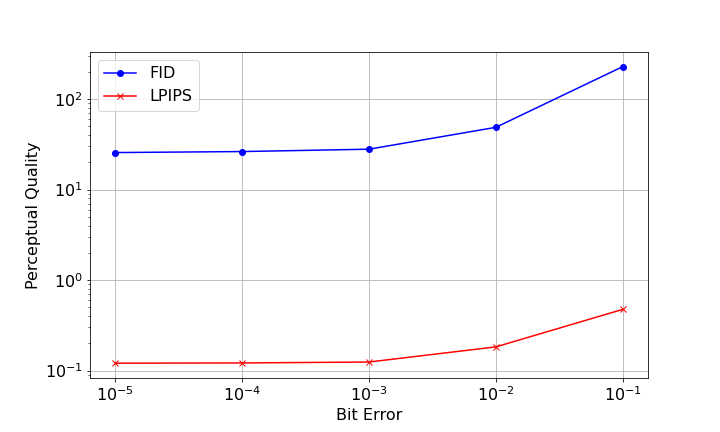}
\caption{Variation of the perceptual quality on the receiver side in terms of FID and LPIPS for different levels of bit error.
}
\label{fig_setup3}
\end{figure} 

In Fig~\ref{fig_setup3}, we present the results for noise setup 3. For this setup, we evaluate the bit error: represented as a flip of bits where fixed-length source coding is used. From Fig.~\ref{fig_setup3}, we see \model perform robustly against a bit error rate of 0.01. In a typical communication system, we can expect a bit error rate of 0.00001. Therefore, we can assert robust performance from our model in practical setups. Note that, now the bandwidth of the system has increased due to fixed-length coding. From Fig.~\ref{fig_setup2} and Fig.~\ref{fig_setup3}, we see a subtle improvement in perceptual quality when we use the block interleaving. This is a result of error distribution by the interleaving.

Finally, we evaluate the packet error without block interleaving.
In fact, random masking PRM with $p=0.35$ is very similar to a communication link with random masking PRM at
$p=0.25$ that undergoes a channel with PER of 10\%. In this set of experiments, we assume that channel noise only 
affects $\mathbf{\hat{z}}$. Therefore, we evaluate the random masking against different metrics as presented in Table~\ref{tab:tablenoise2}. The results indicate that \model demonstrates robustness against packet loss, as evidenced by both FID and LPIPS scores.  Table~\ref{tab:tablenoise} shows the results of 10\% PER under different noising/denoising steps with respect to perceptual metrics and to a downstream task. From this table, it is evident that increasing the number of denoising steps enhances robustness to channel noise, improving both perceptual quality and depth estimation accuracy.
\begin{table}[!t]
\caption{Analysis of model performance against noise. Here, we evaluate the perceptual performance and downstream task performance on the Cityscapes dataset with channel PER = 10\%.  } 
\label{tab:tablenoise}
\centering
\begin{tabular}{|c||c|c|c|c|}
\hline
\textbf{Steps} & \textbf{FID}$\downarrow$  & LPIPS$\downarrow$ & Depth$\downarrow$  \\
\hline
28/32& 30.55  & 0.1560 & 5.99 \\
\hline
50/62& 30.05 & 0.1524&6.08 \\
\hline
94/100& 28.42& 0.1470& 5.97\\
\hline
194/200& \textbf{27.23} & \textbf{0.1438} & \textbf{5.88}\\
\hline

\end{tabular}
\end{table}

\begin{table}[!t]
\caption{Model performance against a noisy channel. Here, we model the noise as packet error rate (PER) with different percentages. We have used the random policy with $p=0.25$ on the cityscape dataset.  } 
\label{tab:tablenoise2}
\centering
\begin{tabular}{|c||c|c|c|c|}
\hline
\textbf{PER} & \textbf{FID}$\downarrow$  & \textbf{LPIPS}$\downarrow$\\
\hline
1\%&  23.18 & 0.1148  \\
\hline
2\%& 23.59 & 0.1179 \\
\hline
3\%& 24.49& 0.1150\\
\hline
5\%& 25.49 & 0.1270 \\
\hline

\end{tabular}
\end{table}

\begin{figure}[h]
\centering
\includegraphics[width=\linewidth]{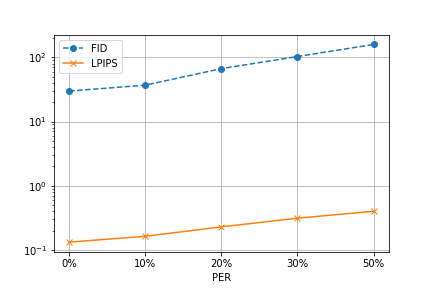}
\caption{Variation of the perceptual quality in a noisy channel. The model is evaluated on the Cityscape dataset with 4/8 noise/denoise steps. 
}
\label{fig_noise1}
\end{figure}

\begin{figure}[h]
\centering
\includegraphics[width=\linewidth]{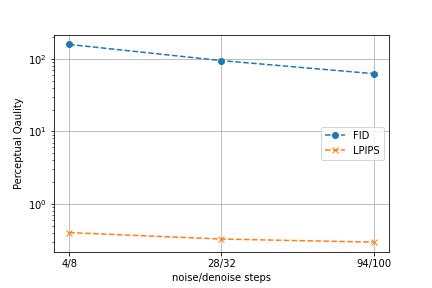}
\caption{Variation of the perceptual quality in a noisy channel according to the noising/denoising steps. Here PER is 50\%.
}
\label{fig_noise2}
\end{figure}

Now we assume a case where channel noise affects both $\mathbf{\hat{z}}$ and the text conditions: $c_T$. As it is not straightforward to simulate an erasure channel with packet loss, we make the following assumptions. Consider wireless transmission is over a WiFi network. We assume a variable payload of around 105 bytes which generates approximately 85 packets per Cityscape image. We assume each packet contains 70 codebook indices or less, or 70 words or less representing the text condition. We assume no packet contains both codebook indices and text conditions. In Fig.~\ref{fig_noise1}, we present the variation of different perceptual metrics under different channel noise conditions. As we expected, the perceptual quality shows a downward trend as the packet error rate displays a rise. The system delivers acceptable perceptual reconstruction up to PER of 20\%. Next, in Fig.~\ref{fig_noise2}, we illustrate the effect of noising and denoising steps under certain PER. For this test, we select the Cityscape dataset and a channel with 50\% of PER and evaluate the perceptual quality of the reconstruction. As we see from the figure, the perceptual quality gradually improves with lower FID and LPIPS scores as we increase the
diffusion steps. Therefore, we observe that increasing the computation steps in diffusion enhances the proposed model's robustness to channel noise.

\begin{figure}[t]
\centering
\includegraphics[height = 1\columnwidth]{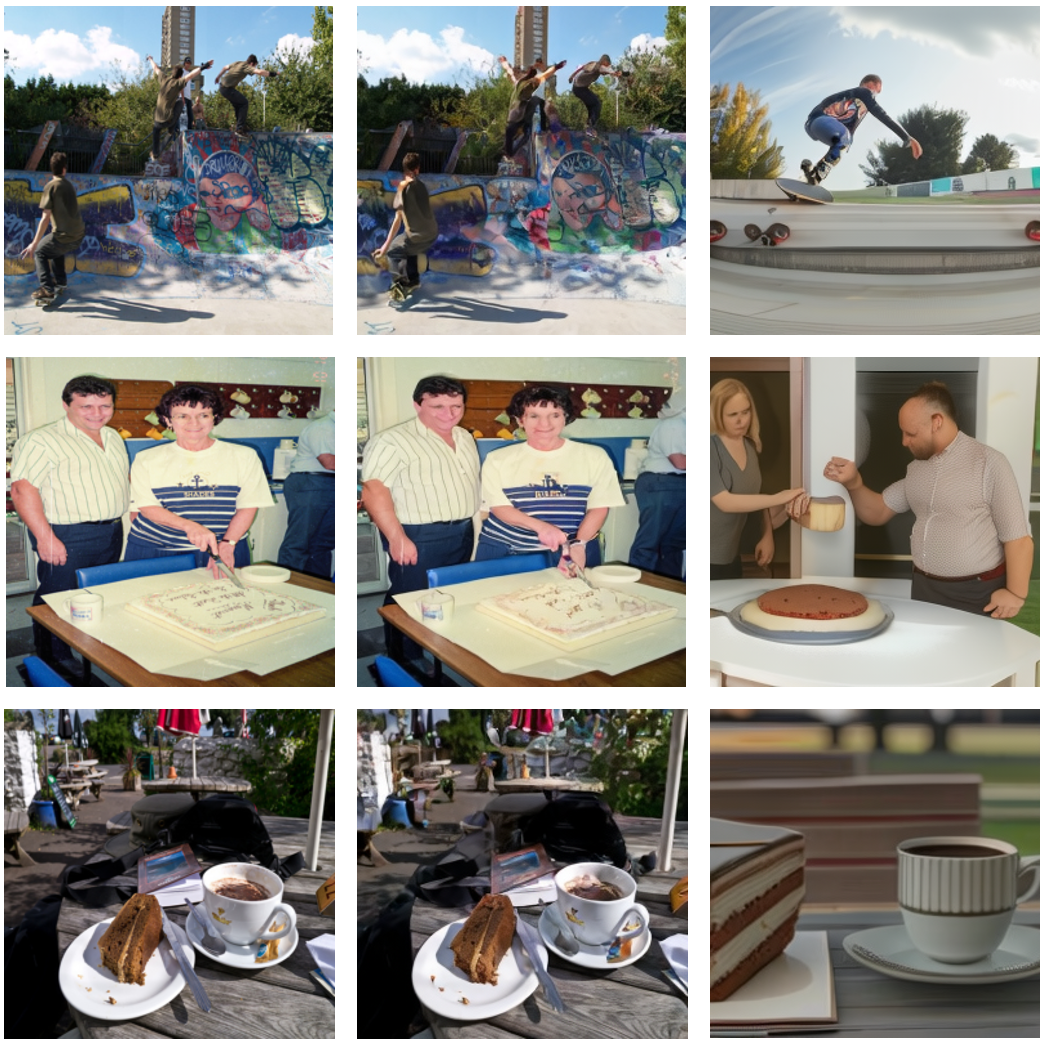}
\caption{Effect of latent mixture integration process. The first column represents original images. The generated image using latent mixture integration is presented in the next column. Finally, the third column is the recovery without latent mixture integration. The reconstructions are generated using the same mode, initial latent, and text conditions. The corresponding text conditions are as follows. T1: ``A young man riding a skateboard into the air. a group of teenagers jumping a ramp on their skateboards. A time lapse image of a guy on a skate board. Group of boys performing skateboard trick on ramp with graffiti on it. some male skateboarders are doing some tricks and grafitti.'' T2: ``A man and woman cut into a big cake . A man and woman standing in front of a cake. A women who is cutting into a cake. two people standing near a table with a cake . A woman cutting into a cake with a man standing behind her.''T3:``A piece of cake and coffee are on an outdoor table. Slice of cake next to large filled cup on wooden table. A dessert and a cup of coffee sit next to a book and a purse. A slice of cake and mug sitting on a wooden table outside. a close up of a slice of cake on a plate.'' }
\label{fig:p1}
\end{figure}

\begin{figure}[h]
\centering
\includegraphics[width=\linewidth]{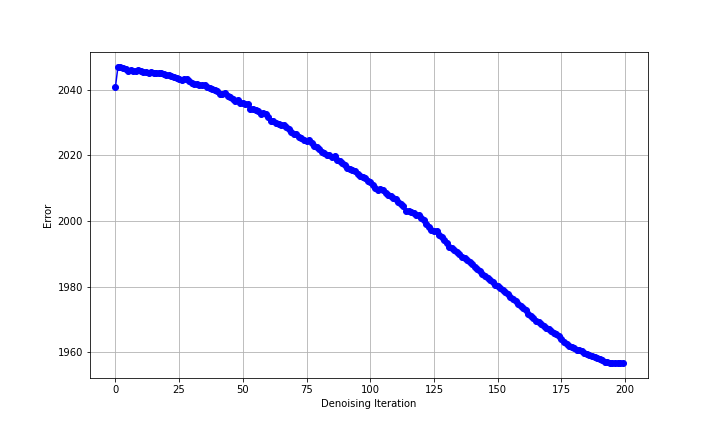}
\caption{Misclassified indices vs denoising iterations. 
}
\label{fig_error}
\end{figure}

\subsection{Convergence Analysis}

In this section, we examine the model convergence at each iteration. First, we analyze the error of the predicted class at each index of $\hat{\mathbf{I}}_i$ and graphically illustrate the error against each denoising iteration in Fig.~\ref{fig_error}. As shown in the results of the figure, the number of class misclassifications decreases as the iteration number increases. However, we see the error flattens after 200 steps. 
We find two reasons behind this observation. First, after obtaining the prediction probabilities, the model randomly samples possible identities from the distribution rather than maximizing the likelihood, and also, as shown in the work~\cite{vq-compression}, many of the latent embeddings are clustered in the embedding space. Hence, gives the freedom of predicting an identifier in the same cluster rather than the exact identifier. We save extensive analysis of convergence as future works. 

\subsection{Analysis of Showtime }
\begin{table}[!h]
\caption{\model showtime. We compare showtime with perceptual metrics and reconstruction steps. All the methods have been tested in the same environment. } 
\label{tab:tabletime}
\centering
\begin{tabular}{|c||c|c|c|c|c|c|}
\hline
Model & Steps ($\tau / T$) & FID$\downarrow$  & LPIPS$\downarrow$& Time (s)$\downarrow$\\
\hline
Diff-GO+& 30/30 &65.81&0.3546&7.2 \\
\hline
\model& 4/8 &29.84&0.1327&1.29\\ 
\hline

\end{tabular}
\end{table}

We extend our analysis to compare the showtime delay of different GO-COM models along with the perceptual quality. While it is a critical metric for assessing a communication system, showtime is also a key consideration in system implementation. For \model, we evaluate the EBM masking policy for the following set of experiments. Note that all the models are tested on the Cityscapes dataset. According to Table~\ref{tab:tabletime}, our \model can perform information reconstruction faster with better reconstruction quality than the previous GO communication methods. Although our \model demonstrates superior performance compared to existing methods, there is a clear need for further reduction in showtime. To achieve this, we advocate for leveraging additional computational power, which is expected to be readily available shortly.

\subsection{Ablation Study}

We perform an ablation study to investigate the effectiveness of our solution. For that, we will ablate the latent mixture integration process to verify the effect of our model. We visually investigate the repercussions on the COCO-Stuff dataset by considering samples illustrated in Fig.~\ref{fig:p1} first column. We use the PDM policy to derive the latent received by the receiver and perform iterative image reconstruction without latent mixture integration. The corresponding results are shown in  Fig.~\ref{fig:p1} third column and the respective text condition is presented in the caption. For
comparison, we present the output of the latent mixture integration by using the same set of text conditions.
The corresponding results are given in Fig.~\ref{fig:p1} second column for comparison. From the above figures, we see a generic image generation without latent mixture   integration 
trying to generate an image that is semantically similar to the text. 
Much encoded information is lost at the receiver. 
We observe that if the receiver's goals include object detection or semantic segmentation, images generated from pure diffusion based on text conditions may carry wrong information. On the other hand, 
latent mixture integration preserves more valuable information and generates an image similar to the original source image.

\section{Conclusion} \label{sec:conclusion}

In this work, we present a new goal-oriented communication 
framework of \model.  Our novel framework,  with generative AI as its backbone, is powered by VQGAN at a higher level to map high-dimensional images to their lower-dimensional latent representations. The latent diffusion model is conditioned on semantic 
descriptive text for image recovery. \model framework achieves higher perceptual quality, higher downstream task performances, faster reconstruction, and robust performance 
against erasure channels for different packet error rates for a reasonable bandwidth. In future works, we plan to integrate diffusion-based GO-COM with neural network pruning to increase the computation efficiency. We will also investigate the distributed training of multi-user GO-COM with advanced foundation models.


\end{document}